\documentclass{article}

     \usepackage[nonatbib, final]{neurips_2019}

\usepackage[utf8]{inputenc} 
\usepackage[T1]{fontenc}    
\usepackage{hyperref}       
\usepackage{url}            
\usepackage{booktabs}       
\usepackage{amsfonts}       
\usepackage{nicefrac}       
\usepackage{microtype}      
\usepackage{float}
\usepackage{graphicx}

\usepackage{cite}

\title{Development of a hand pose recognition system on an embedded computer using CNNs}

\author{%
  Dennis Núñez Fernández
  \\
  Universidad Nacional de Ingeniería 
  \\
  Lima, Peru 
  \\
  \texttt{dnunezf@uni.pe} 
  \\
}

\begin{document}

\maketitle

\begin{abstract}
Demand of hand pose recognition systems are growing in the last years in technologies like human-machine interfaces. This work suggests an approach for hand pose recognition in embedded computers using hand tracking and CNNs. Results show a fast time response with an accuracy of 94.50\% and low power consumption.
\end{abstract}

\section{Introduction}

Hand gesture recognition is one obvious strategy to build user-friendly interfaces between machines and users. In the near future, hand posture recognition technology would allow for the operation of machines through only series of hand postures, eliminating the need for physical contact. However, hand gesture recognition is a difficult problem because occlusions, variations of appearance, etc. Despite these difficulties, several approaches to gesture recognition on images has been proposed \cite{1}.

In recent years, convolutional neural networks (ConvNets) have become the state-of-the-art for object recognition \cite{2}. In spite of the high potential of CNNs in object detection problems \cite{3, 4} and image segmentation \cite{2}, only a few papers report successful results. A recent survey on hand gesture recognition \cite{1} reports only one important work \cite{5}. Some obstacles to wider use of CNNs are high computational costs, lack of sufficiently large datasets, as well as lack of appropriate hand detectors.

\section{Methodology}

The proposed system works with images captured from a CMOS camera and runs on embedded computers without GPU support such as the Raspberry Pi, BeagleBone, Intel Galileo among others. Therefore, the goals of the proposed system are as follows: high accuracy, fast response time and low power consumption. The system was implemented in C++ in order to obtain the best performance.

\begin{figure}[H]
  \centering{\includegraphics[width=90mm]{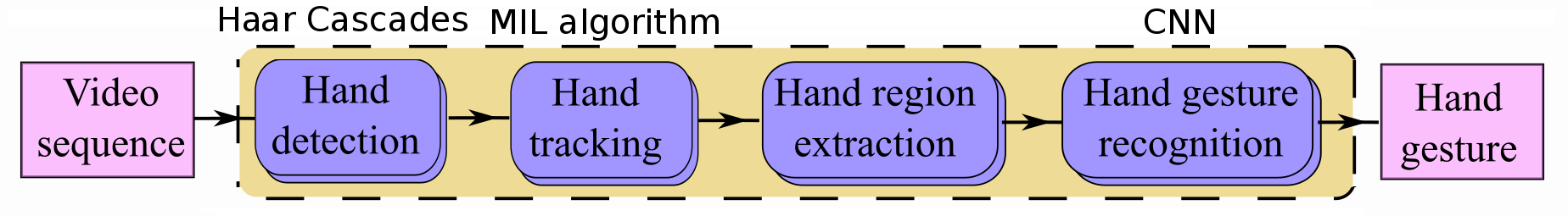}}
  \caption{Diagram for the proposed system}
  \label{diagram}
\end{figure}

Haar cascade classifier allows better detection for objects with static features such as balloons, faces, eyes, etc. But a hand in motion has few static features because shape changes over time. So, this classifier is not suitable to recognize a hand poses in motion. However, its deficiency could be compensated with a hand tracker based on wrist region, which features keep invariant over time. Furthermore, tracking reduces the processing time since it requires less computational resources than detection. We use the MIL (Multiple Instance Learning) tracking algorithm \cite{6}. It avoids the drift problem for a robust tracking and consumes less memory and computational resources than Haar cascade classifier. In addition, due skin color is a powerful feature for fast hand detection, a model in RGB-YCbCr color spaces have been constructed on the basis of a training dataset. Then, the hand region was obtained by thresholding and morphological operations. The dataset for hand gesture classification was taken from AGH University of Science and Technology \cite{7}. It has 73,124 grayscale images of 48x48 pixels divided into 10 hand gestures. The proposed CNN takes as input a binary image of 48x48 pixels. The architecture is: C(5x5)-S(2x2)-C(3x3)-S(2x2)-FC(120)-FC(84)-FC(10), where C: Conv. layer, S: Sub sampling, FC: Full connection. We used Caffe framework \cite{8}.

\section{Results}

\begin{figure}[H]
  \centering
  \begin{minipage}[b]{0.4\textwidth}
    \includegraphics[width=50mm]{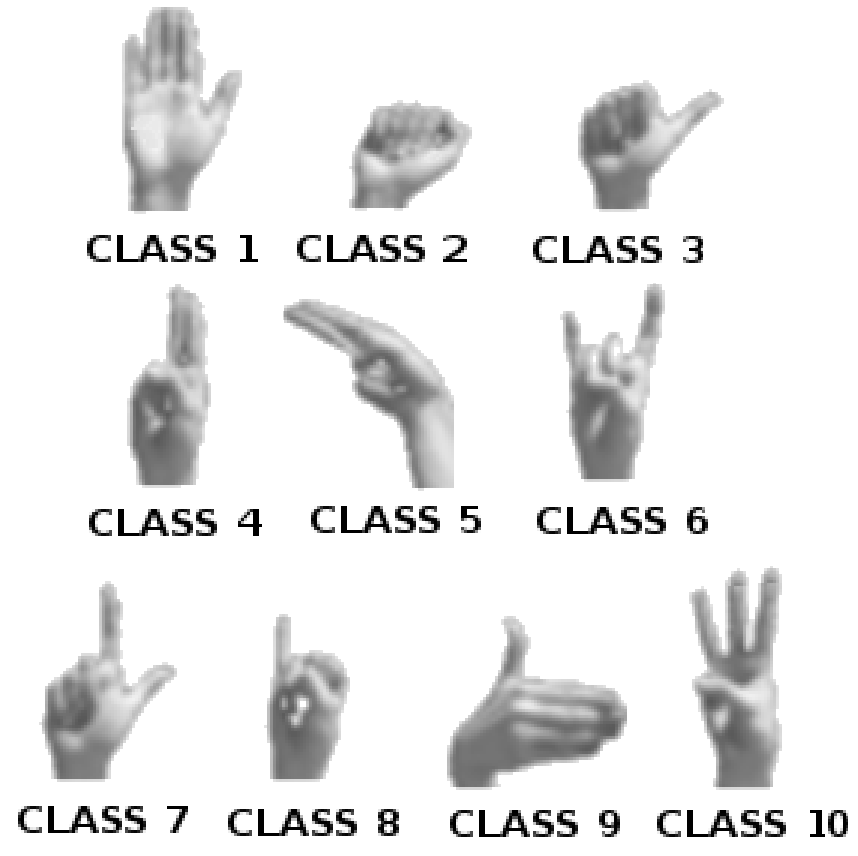}
    \caption{Hand poses}
    \label{hand_poses}
  \end{minipage}
  \begin{minipage}[b]{75mm}
    \includegraphics[width=74mm]{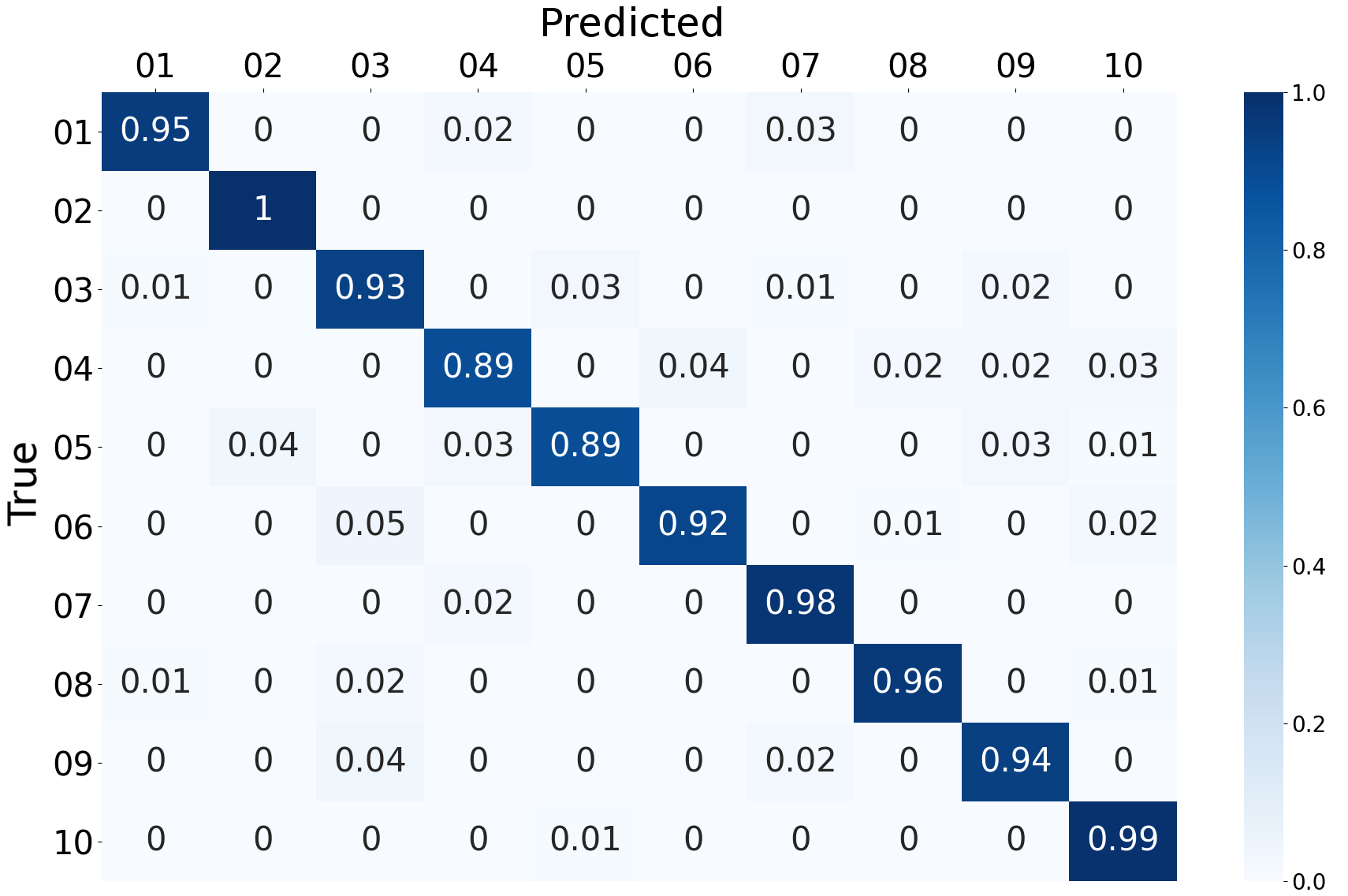}
    \caption{Confusion matrix}
    \label{confusion_matrix}
  \end{minipage}
\end{figure}

The performance of the proposed Convolutional Neural Network for hand poses classification was evaluated using different metrics such as confusion matrix and accuracy. Fig.~\ref{hand_poses} depicts the hand pose for each class in grayscale format. The confusion matrix of our model is shown in Fig.~\ref{confusion_matrix} and discloses which hand poses are misclassified. These errors happen because of similarities between the classes. Furthermore, our architecture shows an outstanding accuracy of 94.50\%.


The implementation of the proposed recognition system on a desktop PC has no issues due to its high computational resources. However, when a recognition system is implemented on embedded computers like the Raspberry Pi 3 we have two major obstacles working against us: limited RAM memory (only 1 GB) and limited processor speed (4 ARM Cortex-A53 @ 1.2 GHz). The Table~\ref{time_comparision_models} shows the performance of CNNs on the Raspberry Pi 3 platform. As you can see, the proposed CNN achieves the fastest response time and the lowest power consumption.

\begin{table}[H]
\small
\caption{Response time and power consumption for different CNNs on a Raspberry Pi 3 using Caffe}
\label{time_comparision_models}
\centering
\begin{tabular}{llllll}
\toprule
Model & Proposed CNN & VGG\_F \cite{9} & NiN \cite{10} & AlexNet \cite{2} & GoogLeNet \cite{11} \\
\midrule
Layers & 9 & 13 & 16 & 11 & 27 \\
Power (W.) & 0.690 & 0.760 & 0.840 & 0.750 & 0.790 \\
Time (s.) & 0.351 & 0.857 & 0.553 & 1.803 & 1.175 \\
\bottomrule
\end{tabular}
\end{table}

\section{Conclusions}

In this work we demonstrated that our system is capable to recognize 10 hand gestures with an accuracy of 94.50\% on images captured from a single RGB camera, and using low power consumption, which is about 0.690 W. In addition, we show that the average time to process each image on the Raspberry Pi 3 is about 351.2 ms. The trained CNN models (Caffe models) as well as a version of the source code are fully available at: \url{https://github.com/dennishnf/cnn-hand-gesture-interface}. The results explained before show that our hand pose recognition system can be used for controlling robots, for virtual reality interaction, for human-machine interfaces among others.

\clearpage

\small

\bibliography{sample}{}

\begin{thebibliography}{10}

\bibitem{4}
Itamar Arel, Derek Rose, and Thomas Karnowski.
\newblock Deep machine learning - a new frontier in artificial intelligence
  research [research frontier].
\newblock {\em IEEE Comp. Int. Mag.}, 5:13--18, 01 2010.

\bibitem{6}
B.~{Babenko}, M.~{Yang}, and S.~{Belongie}.
\newblock Visual tracking with online multiple instance learning.
\newblock In {\em 2009 IEEE Conference on Computer Vision and Pattern
  Recognition}, pages 983--990, June 2009.

\bibitem{9}
K.~Chatfield, K.~Simonyan, A.~Vedaldi, and A.~Zisserman.
\newblock Return of the devil in the details: Delving deep into convolutional
  nets.
\newblock In {\em British Machine Vision Conference}, 2014.

\bibitem{8}
Yangqing Jia, Evan Shelhamer, Jeff Donahue, Sergey Karayev, Jonathan Long, Ross
  Girshick, Sergio Guadarrama, and Trevor Darrell.
\newblock Caffe: Convolutional architecture for fast feature embedding.
\newblock In {\em Proceedings of the 22Nd ACM International Conference on
  Multimedia}, MM '14, pages 675--678, New York, NY, USA, 2014. ACM.

\bibitem{2}
Alex Krizhevsky, Ilya Sutskever, and Geoffrey~E. Hinton.
\newblock Imagenet classification with deep convolutional neural networks.
\newblock In {\em Proceedings of the 25th International Conference on Neural
  Information Processing Systems - Volume 1}, NIPS'12, pages 1097--1105, USA,
  2012. Curran Associates Inc.

\bibitem{3}
Bogdan Kwolek.
\newblock Face detection using convolutional neural networks and gabor filters.
\newblock In W{\l}odzis{\l}aw Duch, Janusz Kacprzyk, Erkki Oja, and S{\l}awomir
  Zadro{\.{z}}ny, editors, {\em Artificial Neural Networks: Biological
  Inspirations -- ICANN 2005}, pages 551--556, Berlin, Heidelberg, 2005.
  Springer Berlin Heidelberg.

\bibitem{10}
Min Lin, Qiang Chen, and Shuicheng Yan.
\newblock Network in network.
\newblock {\em CoRR}, abs/1312.4400, 2013.

\bibitem{7}
Dennis N{\'u}{\~{n}}ez~Fern{\'a}ndez and Bogdan Kwolek.
\newblock Hand posture recognition using convolutional neural network.
\newblock In Marcelo Mendoza and Sergio Velast{\'i}n, editors, {\em Progress in
  Pattern Recognition, Image Analysis, Computer Vision, and Applications},
  pages 441--449, Cham, 2018. Springer International Publishing.

\bibitem{1}
Oyebade Oyedotun and Adnan Khashman.
\newblock Deep learning in vision-based static hand gesture recognition.
\newblock {\em Neural Computing and Applications}, 28, 04 2016.

\bibitem{11}
C.~{Szegedy}, {Wei Liu}, {Yangqing Jia}, P.~{Sermanet}, S.~{Reed},
  D.~{Anguelov}, D.~{Erhan}, V.~{Vanhoucke}, and A.~{Rabinovich}.
\newblock Going deeper with convolutions.
\newblock In {\em 2015 IEEE Conference on Computer Vision and Pattern
  Recognition (CVPR)}, pages 1--9, June 2015.

\bibitem{5}
Jonathan Tompson, Murphy Stein, Yann Lecun, and Ken Perlin.
\newblock Real-time continuous pose recovery of human hands using convolutional
  networks.
\newblock volume~33, 08 2014.

\end{thebibliography}
\bibliographystyle{plain}

\end{document}